\pdfoutput=1
\documentclass[lettersize,journal]{IEEEtran}
\usepackage[T1]{fontenc}    
\usepackage{hyperref}       
\usepackage{url}            
\usepackage{booktabs}       
\usepackage{amsfonts}       
\usepackage{nicefrac}       
\usepackage{microtype}      
\usepackage{xcolor}         
\usepackage{gensymb}
\usepackage{multirow}
\usepackage{makecell}
\usepackage{array}
\usepackage{graphicx}
\usepackage{amsmath}
\usepackage{float}
\usepackage{subfig}

\ifCLASSINFOpdf
\else
\fi
\begin{document}
	%
	\title{Rotation Perturbation Robustness in Point Cloud Analysis: A Perspective of Manifold Distillation}

	
	\author{Xinyu Xu, Huazhen Liu, Feiming Wei, Huilin Xiong, \\ Wenxian Yu,~\IEEEmembership{Senior Member,~IEEE}, and Tao Zhang$^{\ast}$ \thanks{*Corresponding author},~\IEEEmembership{Member,~IEEE},
		\thanks{This paper was supported in part by XXX.}
		\thanks{X. Xu, T. Zhang, F. Wei, H. Xiong, and W. Yu are with the Shanghai Key Laboratory of Intelligent Sensing and Recognition, School of Sensing Science and Engineering, Shanghai Jiao Tong University, Shanghai 200240, China.}
		\thanks{H. Liu is with the Intelligent Photoelectric Sensing Institute, School of Sensing Science and Engineering, Shanghai Jiao Tong University, Shanghai 200240, China.}
	}

	\markboth{Journal of \LaTeX\ Class Files,~Vol.~14, No.~8, August~2015}%
	{Shell \MakeLowercase{\textit{et al.}}: Bare Demo of IEEEtran.cls for IEEE Transactions on Magnetics Journals}
	%



	\IEEEtitleabstractindextext{%
		\begin{abstract}
			Point cloud is often regarded as a discrete sampling of Riemannian manifold and plays a pivotal role in the 3D image interpretation. Particularly, rotation perturbation, an unexpected small change in rotation caused by various factors (like equipment offset, system instability, measurement errors and so on), can easily lead to the inferior results in point cloud learning tasks. However, classical point cloud learning methods are sensitive to rotation perturbation, and the existing networks with rotation robustness also have much room for improvements in terms of performance and noise tolerance. Given these, this paper remodels the point cloud from the perspective of manifold as well as designs a manifold distillation method to achieve the robustness of rotation perturbation without any coordinate transformation. In brief, during the training phase, we introduce a teacher network to learn the rotation robustness information and transfer this information to the student network through online distillation. In the inference phase, the student network directly utilizes the original 3D coordinate information to achieve the robustness of rotation perturbation. Experiments carried out on four different datasets verify the effectiveness of our method. Averagely, on the Modelnet40 and ScanobjectNN classification datasets with random rotation perturbations, our classification accuracy has respectively improved by 4.92\% and 4.41\%, compared to popular rotation-robust networks; on the ShapeNet and S3DIS segmentation datasets, compared to the rotation-robust networks, the improvements of mIoU are 7.36\% and 4.82\%, respectively. Besides, from the experimental results, the proposed algorithm also shows excellent performance in resisting noise and outliers. 
		\end{abstract}
		
		}

	\maketitle

	\IEEEdisplaynontitleabstractindextext

	%
	\IEEEpeerreviewmaketitle

	\section{Introduction}
	%
	%
	%
	%
	\IEEEPARstart{P}{oint} cloud, collections of densely distributed unordered points in three-dimensional space, is often regarded as two-dimensional manifolds embedded in three-dimensional space, offering rich information for numerous tasks such as 3D reconstruction \cite{dai2021indoor}, object recognition \cite{qi2017pointnet}, and scene understanding \cite{zhao2022real}. 
	
\begin{figure}
		\centering
		\includegraphics[width=3in]{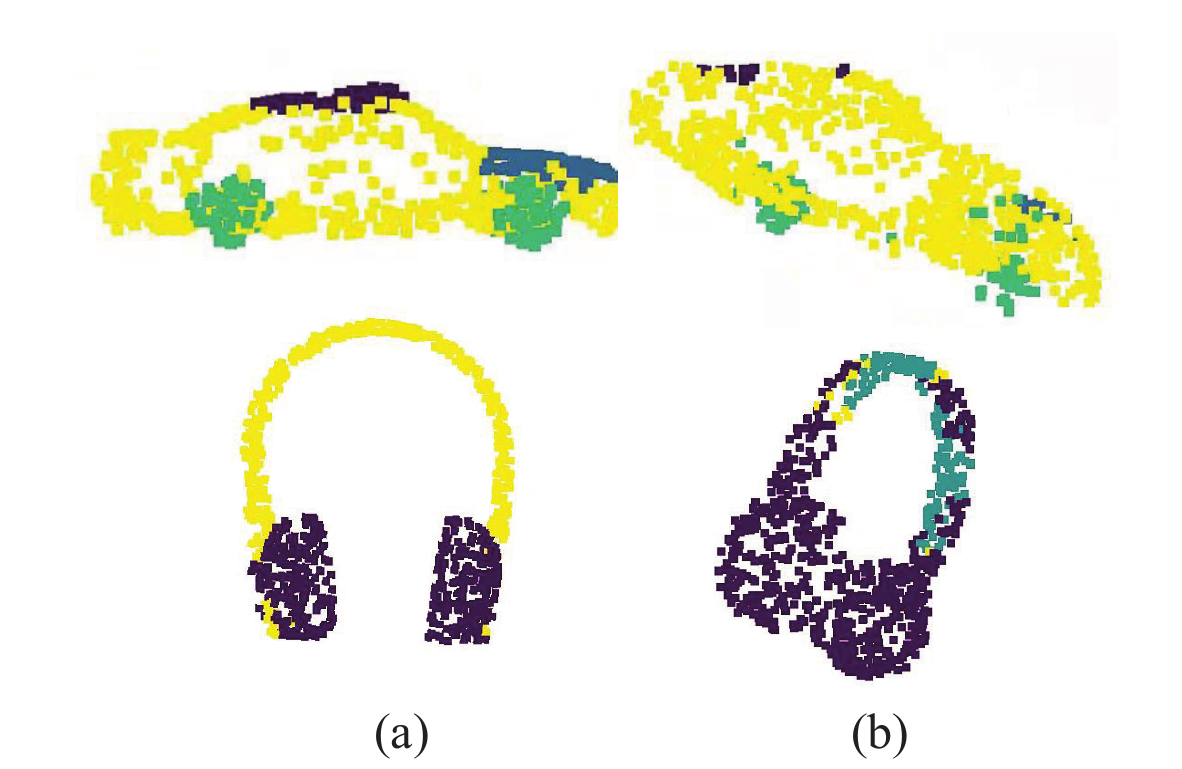}
		\caption{Different segmentation results of PointNet++. (a) Original segmentation results; (b) The segmentation results when the axis is rotated 30\degree.}
		\label{visual_pointnet++}
	\end{figure}

	In recent years, deep learning has achieved remarkable accomplishments in the field of point cloud interpretation \cite{sun2021quadratic,zhao2022jsnet++,wang2023domain}. This technological breakthrough is primarily attributed to the efficient learning and representation capabilities of deep neural networks that empower them to discover and harness the abundant geometric information and spatial structures embedded within point cloud data.
	To be specific, existing algorithms typically perceive point cloud as two-dimensional manifolds embedded in a vertically oriented three-dimensional space, leveraging hierarchical structures to extract intrinsic feature information of point cloud so as to realize high-precision classification and segmentation tasks \cite{yin2023dcnet,phan2018dgcnn,uy2019revisiting}. Nevertheless, in real-world point cloud acquisition processes, apart from random errors such as noise and outliers \cite{rakotosaona2020pointcleannet}, systematic errors including rotation perturbation caused by equipment offset or system instability \cite{zhang2019rotation} also pose a significant challenge to the robustness and stability of point cloud analysis. Taking the classical network PointNet++ \cite{qi2017pointnet++} as an example, one can see from Fig. \ref{visual_pointnet++} that when unforeseen deviations occur in the test data, the segmentation results are quite different compared with the original segmentation results.
	
	In order to solve this complex issue, data augmentation is usually seen as a direct solution, but maintaining efficient and effective information learning capabilities is difficult due to the redundant calculations and imbalanced data distribution \cite{10182102}. Furthermore, some cutting-edge researches also focus on achieving rotation robustness within the architectures of networks, which can be mainly classified to three categories. The first one is committed to reconstructing the local coordinate system of manifold to accurately extract geometric features such as distances and angles between adjacent points \cite{zhang2022riconv++}. The second one performs regularizations on the manifold coordinates to correct the pose and embed them into the Euclidean space from different perspectives \cite{li2021closer}. The last one maps the manifold coordinates from the Euclidean space to the spherical space, and then slides a spherical voxel convolution kernel in the space to achieve rotation robustness \cite{you2021prin}. Nevertheless, these methods, while enhancing the model's adaptability to rotation perturbation, often overemphasize the extraction of rotation-robust information, compromising generalization capabilities in the presence of random errors (such as noise and outliers) \cite{kim2020rotation}. Besides, they typically introduce extra computational burdens during the inference process as well, owing to the required coordinate transformation. Therefore, to achieve robustness of rotation perturbation efficiently while maintaining the ability to resist random errors still remains an unresolved issue. For this goal, we propose a novel augmentation strategy based on manifold distillation, using a teacher-student architecture with several designed models to achieve the robustness against rotation perturbation and random errors without relying on complex coordinate transformation and increasing the number of inference parameters during inference phase. The contributions of this paper are summarized below:
	\begin{itemize}
		
		\item{We remodel point cloud from the perspective of manifold and construct a hierarchical network architecture to learn the mappings between different manifolds while extracting the structural information of manifolds with a graph representation method.}
		\item{A novel augmentation approach based on manifold distillation is introduced. During the training phase, a teacher-student architecture is utilized to align the feature information of the student network with the rotation-robust manifold provided by the teacher network. In the inference phase, the teacher network is removed, and the student network directly leverages the original coordinates to achieve the robustness against the rotation perturbation.}
		\item{To avoid the degradation and distortion of information caused by direct manifold alignment, we use low-rank decomposition to extract manifold structural components, and further utilize the local Euclidean space properties of manifold to design an alignment model. Meanwhile, to eliminate the additional computational overhead introduced by above processes, a reparameterization technique is also proposed.}
		\item{Experiments conducted on two classification datasets ModelNet40, ScanobjectNN and two segmentation datasets PartNet, S3DIS show that our method not only realizes the robustness against the rotation perturbation, but also successfully resists noise and outliers.}
	\end{itemize}
	
	The remainder of this paper is organized as follows. In Section \ref{2}, we discuss the existing works about the point cloud-related tasks. Section \ref{3} introduces how to realize anti-perturbation learning in primitive Euclidean space by distillation learning. Section \ref{4} shows experimental results. The limitations of our method are presented in Section \ref{5} and conclusions are shown in Section \ref{6}.
	
	\section{Related Works}
	\label{2}
	
	\subsection{Rotation-sensitive Learning on 3D Point Cloud}
	With the continuous breakthroughs in deep learning, the point cloud learning methods based on 3D coordinates gradually comes into the spotlight, which has also fueled the development of autonomous driving, robot navigation, and other fields. As a pioneering work, PointNet \cite{qi2017pointnet} extracted global features from the unordered original points, achieving the performance of universal approximation while ensuring permutation invariance. In \cite{qi2017pointnet++}, PointNet++ was proposed that applied PointNet recursively on a nested partitioning of the input points to learn deep features more efficiently and robustly. DGCNN \cite{phan2018dgcnn}, as a multi-layer convolutional neural network based on labeled directed graphs, adapted to the dynamic structure of local regions inside graphs through flexibly designed convolution kernels, demonstrating superior point cloud identification results. Su $et\ al.$ \cite{su2023weakly} put forward a multi-prototype classifier MulPro according to the characteristics of large intra-class variation and the clear subclass structure in 3D point cloud segmentation, reducing the difficulty of representation learning when there is only sparse labeled data. Although these networks have achieved impressive results in various tasks, when facing rotation perturbation, the deviation between the feature patterns learned by networks and the input data still easily intensifies the complexity of tasks, incurring a significant decline in performance  \cite{xu2021sgmnet}.
	
	\subsection{Rotation-robust Learning on 3D Point Cloud}
	To enhance the robustness of models against the potential rotation perturbation, numerous researches are directed towards altering the embedding methods of manifolds through coordinate transformations for extracting rotation-robust features. As a groundbreaking contribution, Zhang $et\ al.$ \cite{zhang2019rotation} leveraged low-level rotation-robust geometric features such as distances and angles to design convolution operators for point cloud learning, successfully realizing the robustness against 6-DoF transformations such as translation and rotation. Afterwards, Zhang $et\ al.$ further proposed Riconv++ \cite{zhang2022riconv++}, wherein a more concise and efficient convolutional operator was adopted to enhance the discrimination of features, achieving more accurate recognition results. Li $et\ al.$ \cite{li2021closer} designed a simple pose selector module that effectively eliminated ambiguity in principal component analysis (PCA) via conducting an in-depth analysis of the canonical poses, resulting in a significant performance enhancement. SPRIN \cite{you2021prin} utilized density aware adaptive sampling to construct spherical signals, effectively addressing potential distorted point distributions in spherical space. Additionally, this framework also introduced spherical voxel convolution and point resampling strategies to extract rotation-invariant features from each point. Although these methods perform well in resisting the rotation perturbation, there is still room for further improvements in terms of network performance and noise tolerance \cite{zhao2022rotation}.
	
	\subsection{Distillation Learning}
	Knowledge distillation, originally proposed by Hinton \cite{hinton2015distilling} in 2015 and applied to recognition tasks, utilizes a teacher-student architecture to train networks and achieves higher accuracy. It can be primarily categorized into offline distillation \cite{passalis2018learning}, online distillation \cite{zhang2021student}, and self-supervised distillation \cite{chen2023sd}. In offline distillation, the teacher network and the student network exist independently, with the student network learning by imitating the output of teacher network. In contrast, online distillation allows both networks to learn from each other during the training process, achieving a mutually beneficial performance. Self-supervised distillation does not require an external teacher network, instead generating supervisory information through its own prediction results. In this paper, our network is proposed based on the online distillation, which will introduce a rotation-robust teacher branch only during the training phase to collaborate with the student network. This strategy allows the network to learn the geometric features from original 3D data and the invariant information from transformed coordinates. Consequently, the proposed network can achieve the robustness against the rotation perturbation without any redundant coordinate transformations in the inference process.
	
	\section{Methodology}
	\label{3}
	\subsection{Preliminaries}
	\label{3.1}
	\textbf{Manifold and Point Cloud}\quad A manifold is a topological space $\mathcal{M}$ covered by a family of open sets: $\mathcal{M} = { \cup_\alpha} {U_\alpha }$. For every open set ${U_\alpha }$, there exists a homeomorphism  coordinate mapping ${\varphi _\alpha }:{U_\alpha } \to {\mathbb{R}^n}$. ${U_\alpha }$ and ${\varphi _\alpha }$ together constitute a local chart, and all local charts form an atlas $\Lambda  = \{ ({U_\alpha },{\varphi _\alpha })\}$ of manifold. Through $\Lambda$, Euclidean space coordinates can be assigned to each point on $\mathcal{M}$.
	
	Point cloud is discrete and disordered, and can be regarded as the discrete sampling of $\mathcal{M}$ \cite{hu2021dynamic}. The point cloud $\mathcal{P}$ is often directly embedded in a low-dimensional Euclidean space ${\mathbb{R}^3}$ or ${\mathbb{R}^6}$, which is referred to the background space of point cloud. The coordinates on the background space reflect the actual pose of point cloud as well as their corresponding manifold.  The sampling result of  $\mathcal{P}$ in the open set ${U_\alpha }$ is denoted as ${P_\alpha }$, referred to a patch. When the manifold $\mathcal{M}$ is mapped to the target manifold $\mathcal{M'}$, the point cloud on $\mathcal{M}$ undergoes a corresponding transformation, and its image is equivalent to a resampling of $\mathcal{M'}$.
	
\begin{figure}
		\centering
		\includegraphics[width=3.5in]{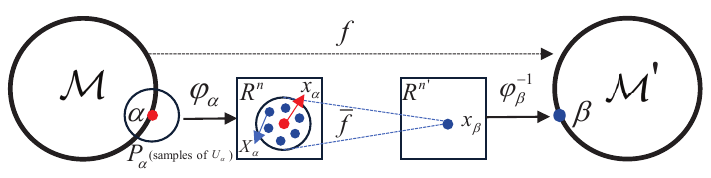}
		\caption{The procedure of manifold mapping. 
			$\alpha$ is a point on the manifold $\mathcal{M}$, open set ${U_\alpha }$ is the neighborhood of $\alpha$, and ${P_\alpha }$ is the discrete sampling of ${U_\alpha }$. A coordinate mapping ${\varphi _\alpha }$ maps $P_{\alpha}$ to $X_{\alpha}$, which is located in local Euclidean space $R^n$, while the image of $\alpha$ is $x_{\alpha}$. $\bar f$, a mapping from $R^n$ to $R^{n'}$, maps $X_{\alpha}$ to $x_{\beta}$, which is the image by applying mapping ${\varphi _\beta }$ on $\beta$ of the manifold $\mathcal{M'}$.  This process equals to the manifold mapping $f$ from $\mathcal{M}$ to $\mathcal{M'}$.}
		\label{math model}
	\end{figure}

	\textbf{Manifold Mapping}\quad Manifold mapping maps $\mathcal{M}$ to $\mathcal{M'}$, which can be expressed as
	\begin{equation}
		\mathcal{M'} = f \circ \mathcal{M},
	\end{equation}
	where $f$ represents the mainfold mapping function. Specifically, $f$ consists of four steps. Firstly, $f$ establishes a local chart $({U_\alpha },{\varphi _\alpha })$ for any point $\alpha$ on $\mathcal{M}$. The coordinate map ${\varphi _\alpha }$ then converts $\alpha$ to ${x_\alpha }$ in Euclidean space, i.e., ${x_\alpha } = {\varphi _\alpha } \circ \alpha $. Next, $f$ learns a mapping $\bar f$ that maps the open set ${U_\alpha }$ to a point ${x_{\beta}}$ in a high-dimensional feature space. $\varphi _{\beta}^{ - 1}$ ultimately maps ${x_{\beta}}$ to $\beta$ in $\mathcal{M'}$. The whole process is shown in Fig. \ref{math model} and can be mathematically described as
	\begin{equation}
		\beta = f \circ \alpha  = \varphi _{\beta}^{ - 1} \circ \bar f \circ {\varphi _\alpha } \circ {U_\alpha } = \varphi _{\beta}^{ - 1} \circ {x_{\beta}},
	\end{equation}
	where $\varphi_{\beta}^{-1}$ is the inverse mapping of coordinate mapping $\varphi _{\beta}$. In practical computations, when point cloud undergoes a manifold mapping, it is divided into independent sets of patches $\{ {P_\alpha }\}$ that serve as discrete samples for ${\{U_\alpha }\}$. Each ${P_\alpha } $ consists of $k$ neighboring nodes ${\alpha _1},{\alpha _2},...,{\alpha _k}$ around the central point ${\alpha}$, and is mapped to ${X_\alpha } = \{ {x_{{\alpha _1}}},{x_{{\alpha _2}}}, \ldots ,{x_{{\alpha _k}}}\} $ through ${\varphi _\alpha }$. Since the point cloud is embedded in Euclidean space, after passing through the feature mapping $\bar f$, we directly set $\varphi _{\beta}$ as the identity mapping during the forward inference process without introducing any new calculation. 

	\subsection{Overall Framework}
	\begin{figure*}[ht]
		\begin{equation}
			\begin{array}{l}
				x_{{\alpha _i}}^T = (||{\alpha _i} - \alpha ||,\angle (LRA[{\alpha _i}],\overrightarrow {{\alpha _i}\alpha } ),\angle (LRA[\alpha ],\overrightarrow {{\alpha _i}\alpha } ), \pm \angle (LRA[{\alpha _i}],\overrightarrow {{\alpha _i}\alpha } ),\\[1mm]
				\angle (\overrightarrow {{\alpha _{i + 1}}\alpha } ,\overrightarrow {{\alpha _i}{\alpha _{i + 1}}} ),\angle (LRA[{\alpha _i}],\overrightarrow {{\alpha _i}{\alpha _{i + 1}}} ),\angle (LRA[{\alpha _{i + 1}}],\overrightarrow {{\alpha _i}{\alpha _{i + 1}}} ), \pm \angle (LRA[{\alpha _i}],LRA[{\alpha _{i + 1}}]),
			\end{array}
			\label{components}
		\end{equation}
		{\noindent}	 \rule[-10pt]{17.5cm}{0.05em}
	\end{figure*}
	
\begin{figure*}
		\centering
		\includegraphics[width=7in]{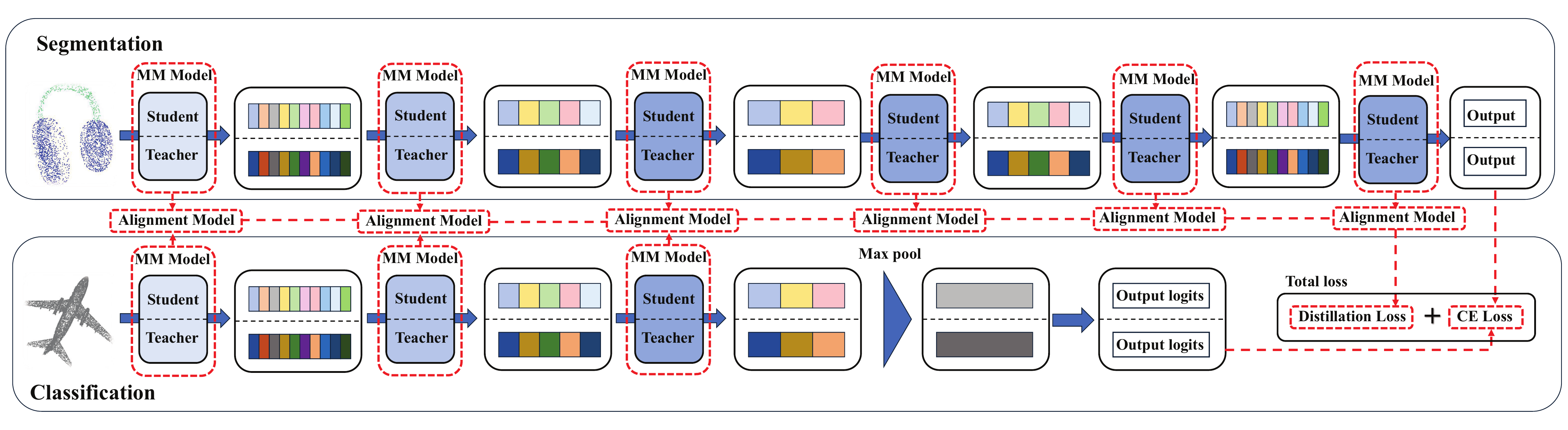}
		\caption{Overall framework of our network.}
		\label{overall framework}
	\end{figure*}

	The essence of the network training process based on point cloud data lies in deeply learning and mastering the complex and precise mapping relationships between different manifold spaces. In this process, the point cloud data can be deemed as a collection of sampling points at corresponding locations on these manifolds. As shown in Fig. \ref{overall framework}, we design a hierarchical approach aimed at efficiently extracting key features from the point cloud data with different receptive fields. In detail, during the downsampling process, we employ multiple manifold mapping models (MM models) to precisely achieve the mapping between manifold structures and aggregated information, while progressively reducing the patches, enabling the network to capture more global information.
	During the upsampling process, we select the farthest points from the downsampling process as new sampling points to ensure the representativeness and uniformity of spatial distribution. The values of these new sampling points are determined by inverse distance weighting. Specifically, different weights are assigned based on the distance from each point to the new sampling point, with closer points having a greater influence. By doing so, we can comprehensively consider the information of surrounding points, making the values of new sampling points more accurate and reasonable. At the same time, to achieve the robustness of rotation perturbation, we design a teacher-student architecture, which aligns the feature information of student network to the rotation-robust features provided by the teacher network through the alignment module. The teacher branch is the rotation-robust network Riconv++ \cite{zhang2022riconv++}, which is insensitive to pose variations in the point cloud, while the student branch extracts and learns local features of point cloud through a graph-based structure, and receives supervision from the teacher branch. During inference phase, the teacher network is discarded and the student network directly uses the original 3D coordinates to obtain perturbation robustness.
	
	\subsection{MM Model}
	\label{3.3}
	
	\begin{figure*}
		\centering
		\includegraphics[width=7in]{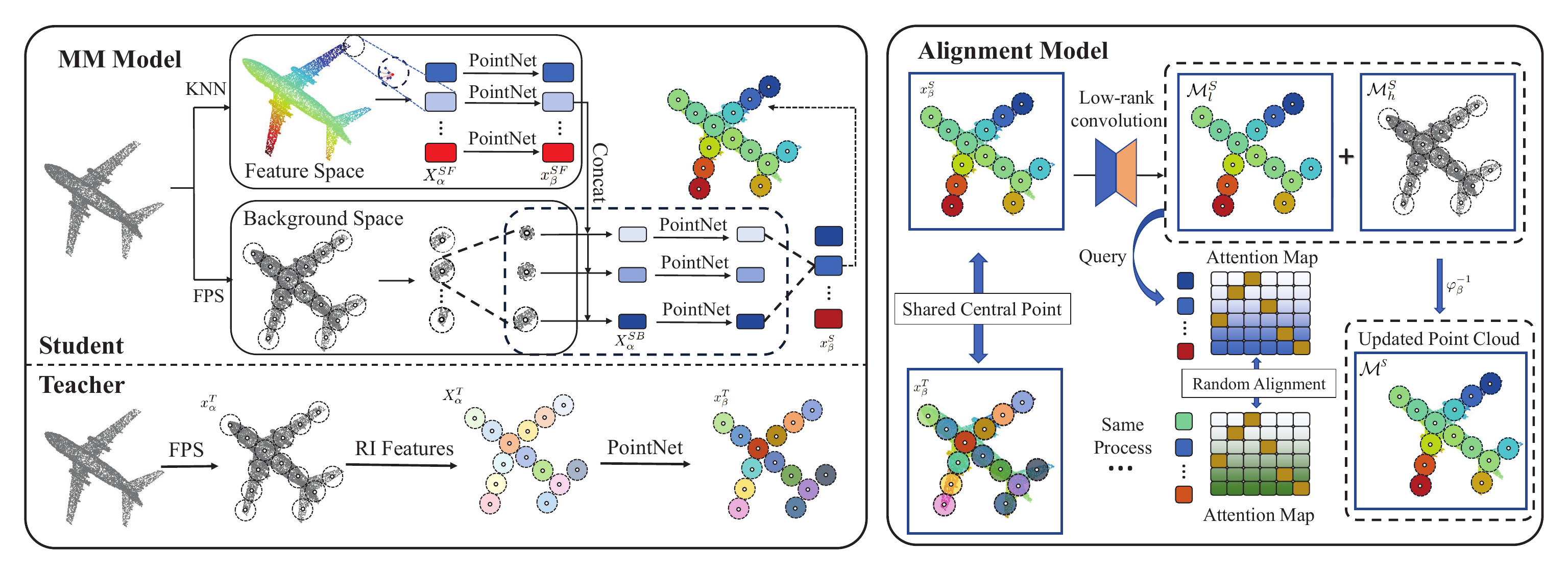}
		\caption{The architectures of MM model and alignment model.}
		\label{Model}
	\end{figure*}
	\textbf{Teacher Network}\quad The mapping of teacher network should be independent of the pose and direction of manifold embedding. To achieve this, a special coordinate mapping ${\varphi _\alpha }$ needs to be established for each patch. The center point $\alpha$ of each patch is obtained by the farthest point sampling in the background space, while the neighboring nodes $P_\alpha ^T$ are obtained through k-nearest neighbors \cite{cover1967nearest}. Then, ${\varphi _\alpha }$ maps $P_\alpha ^T$ to $X_\alpha ^T = \{ x_\alpha ^T\} $, and its coordinate components can be written as Eq. (\ref{components}), where LRA stands for local reference axis, and $ \pm $ depends on the spatial relationship between $\alpha _{i}$ and $\alpha$, as detailed in \cite{zhang2022riconv++}. The first component of $x_{{\alpha _i}}^T$ represents the distance, while the remaining seven components represent the angles, which are rotation-robust features. After obtaining $X_{{\alpha}}^T$, we use PointNet for the manifold mapping $\bar f$ to get $x_{{\beta}}^T$.

	\textbf{Student Network}\quad Unlike the teacher network, the coordinate mapping $\varphi _{\alpha}$ of student network is an identity mapping, i.e., ${x_\alpha }={\varphi _\alpha } \circ {\alpha }={\alpha }$, ${X_\alpha }={\varphi _\alpha } \circ {P_\alpha }={P_\alpha }$. To capture the structural relationships within point cloud, we build graph representation $G$ for each patch. To be specific, the vertices of graph are the central point ${x_\alpha }$ and its neighborhood ${X_\alpha } = \{ {x_{{\alpha _1}}},{x_{{\alpha _2}}}, \ldots ,{x_{{\alpha _k}}}\} $, while the edges are the vectors from each point in ${X_\alpha }$ to ${x_\alpha }$. The graph representation $G$ fuses both of them, which can be expressed as
	\begin{equation}
		\label{eq3}
		G({X_\alpha }) = \{ {x_\alpha } \oplus ({x_{{\alpha _i}}} - {x_\alpha })\},
	\end{equation}
	where $\oplus$ represents the concatenation of dimensions. Then, we establish graph-based mapping ${f_g}$ to realize $\bar f$.  ${f_g}$ learns the advanced semantic features of each patch through PointNet, which achieves feature mapping using a shared-weight multi-layer perception (MLP) and max-pooling to aggregate semantic features from unordered point clouds without extra coordinate preprocessing.
	
	Moreover, in order to extract point cloud features more comprehensively, the student network in MM model extracts the information from both the background space and the feature space. Note that, for the feature space, the semantic information of point cloud manifold is relatively abundant, so we directly use each point in the feature space as the center point of patch and then utilize the k-nearest neighbors algorithm to sample the manifold to obtain $\{ X_\alpha ^{SF}\} $. After establishing the graph representation for $\{ X_\alpha ^{SF}\} $ by Eq. (\ref{eq3}), the graph-based mapping ${f_g}$ further maps it to a new feature space and obtains $\{ x_{\beta}^{SF}\}$. 
	
	In the background space, the point cloud is represented in intuitive coordinates, allowing the use of farthest point sampling to ensure uniform distribution of patch centers on the manifold surface. However, the limited number of channels results in a low-level representation with insufficient semantic details. Thus, adding structural information is necessary to enrich the representation. Specifically, taking the center point of each patch as the sphere center, we adopt the spherical query strategy to find neighboring nodes $X_\alpha ^{SB}$ and set different query radius to artificially divide $X_\alpha ^{SB}$ into multiple sub-regions $\{ X_\alpha ^{S{B_i}}\} $, thereby obtaining a hierarchical graph structure $G(X_\alpha ^{SB})$=$\mathop  \oplus \limits_i G(X_\alpha ^{S{B_i}})$ within the patch. To process this structure, we employ graphwise separable mapping (GSM) for information extraction, which can be described as
	\begin{equation}
		\begin{aligned}
			x_\beta^S &= f_g^{SB}[G(X_\alpha ^{SB}) \oplus \{ x_{\beta}^{SF}\} _\alpha ^{SB}] \\
			&= \mathop  \oplus \limits_i f_g^{SBi}[G(X_\alpha ^{S{B_i}}) \oplus \{ x_{\beta}^{SF}\} _\alpha ^{S{B_i}}],
		\end{aligned}
	\end{equation}
	where $\{ x_{\beta}^{SF}\} _\alpha ^{SB}$ and $\{ x_{\beta}^{SF}\} _\alpha ^{S{B_i}}$ are the sets of the corresponding points of $X_\alpha ^{SB}$ and $X_\alpha ^{S{B_i}}$ in the feature space, respectively. GSM reduces the repeated usage of parameters and conducts targeted learning for the graphs of different sizes. The specific architecture of MM model is shown in Fig. \ref{Model}.
	
	Finally, at the end of MM model, the coordinate mapping $\varphi _{\beta}^{ - 1}$ could be set as identity mapping, in which situation, we do not consider the alignment relationship between the teacher and student networks. 
	
	\subsection{Alignment Model}
	In order to achieve perturbation robustness, the key point is that the manifold mapped by the student network is as close as possible to the rotation-robust manifold provided by the teacher network. However, due to the heterogeneous characteristics of teacher-student architecture, there are significant differences in the structures of network. Therefore, directly aligning the manifolds of student and teacher networks may lead to the loss and distortion of information, further resulting in a decrease in network performance, which can be observed from our subsequent ablation experiments. To deal with this issue, we further design a new alignment model based on the trainable $\varphi _{\beta}^{ - 1}$ for distillation learning, after getting $x_{\beta}^T$ and $x_{\beta}^S$ from the MM model.

	\textbf{Manifold Distillation for Alignment} Due to the differences in structures between the two manifolds, we prefer to maintain a similar spatial information in the local manifolds instead of directly aligning the two manifolds. For the output $x_{\beta}$ (including student and teacher networks) of each layer, we construct the coordinate mapping by splitting it into low frequency parts $x_{\beta, l}$ and high frequency parts $x_{\beta, h}$. So, we can remodel the manifold $x_{\beta}$ as follows
	\begin{equation}
		x_{\beta} = x_{\beta, l} + x_{\beta, h}.
	\end{equation}
	
	Besides, we use the manifolds $\mathcal{M}_h$ and $\mathcal{M}_l$ to represent the constructed manifolds from $x_{\beta, l}$ and $x_{\beta, h}$, respectively. The former retains the main backbone and structure of manifold, and the latter retains the characteristics of manifold. This operation usually requires several iterations of high-dimensional parameters, which is computationally expensive, but can be modeled using a differential neural networks \cite{lee2012smooth}, as shown in Fig. \ref{Model}. Extraction of low-frequency manifold $\mathcal{M}_l$ is conducted by a pair of MLP layers based on the low-rank decomposition \cite{LoRA}, which can be expressed as
	\begin{equation}
		x^T_{\beta, l} = B^TA^TX^T_{\beta}, \\[1mm]
		x^S_{\beta, l} = B^SA^SX^S_{\beta},
	\end{equation}
	where $A^T, A^S  \in \mathbb{R}^{C\times r}$, $B^T,B^S  \in \mathbb{R}^{r\times C}$ are trainable parameters, $C$ is the dimension of $\mathcal{M}$, $r$ is the dimension or rank of $\mathcal{M}_l$. $X^S_{\beta}$ and $X^T_{\beta}$ are patches around $x^S_{\beta}$ and $x^T_{\beta}$ respectively, which can be regarded as local Euclidean space due to the essential property of manifold. According to Whitney embedding theorem \cite{lee2012smooth}, we set the number of $r$ to be less than half the number of input channels $C$ to ensure that $\mathcal{M}_l$ can be re-embedded once aligned.
	
	Next, we consider the alignment between $x^T_{\beta, l}$ and $x^S_{\beta, l}$. Drawing upon the properties of local Euclidean space and inspired by the self-attention mechanism \cite{Attention}, we compute the attention map between patches by multiplying each patch with its transpose. We then use the Kullback-Leibler divergence to perform the alignment. The specific formula can be found in Section III.E. Moreover, the alignment model necessitates that both the teacher network and the student network share the information about the center points of patches to ensure a one-to-one correspondence of characteristic components for each patch. 
	
	\textbf{Reparameterization} The new branch inevitably causes the increase of memory and time consumption. To save the calculation cost during inference, the simply reparameterization technology is further used to fuse the difference process into one branch, i.e., 
	\begin{equation}
		\beta^S = \varphi _{\beta}^{ - 1}(x_{\beta})= (I-B^SA^S+Q^SB^SA^S)X^S_{\beta},
	\end{equation}
	where $I$ represents the identity matrix, $Q^S$ means the student query matrix. As a result, in the inference phase, we could obtain $\beta^S$ by $x_{\beta}$ and trainable parameters instead of relying on frequency estimation.
	
	\subsection{Training Loss}
	\begin{figure*}
		\begin{equation}
			\begin{array}{c}
				{{\cal L}_{KL}} = \frac{{{{\cal T}^2}}}{C}[{\lambda _1}KL(Attn(\beta ^T _l) ||Attn(\beta ^S _l)) + {\lambda _2}KL(Attn(\beta ^S _l)||Attn(\beta ^T _l))],\\[2mm]
				Attn(\beta _l) = Q\beta _ltranspose(Q\beta _l),
			\end{array}
			\label{loss}
		\end{equation}
		{\noindent}	 \rule[-10pt]{17.5cm}{0.05em}
	\end{figure*}
	\textbf{Distillation Loss} The distillation loss is mainly composed of two parts. On one hand, we use the Kullback-Leibler divergence to achieve the alignment in the alignment model, which is expressed in Eq. (\ref{loss}), where $KL$ means the Kullback-Leibler divergence, $\cal T$ is the distillation temperature, $\lambda _1$ and $\lambda _2$ are hyperparameters, $Q$ represents the query matrix, equaling to key matrix. In practical computations, owing to the high time complexity associated with the calculation of attention maps, we further adopt a random sampling approach to extract attention scores from the attention map and perform alignment on them.
	
	On the other hand, in order to prevent the degeneration of manifold structure in $\mathcal{M}_l$, we restrict $\mathcal{M}_h$ by Normalized Mutual Information (NMI) to prevent from containing too much manifold information. The NMI loss can be written as
	\begin{equation}
		{\cal L}_{NMI}(x_{\beta, h}, x_\beta) = \frac{I(x_{\beta, h}, x_\beta)}{\sqrt{H(x_{\beta, h})H(x_{\beta})}},
		\label{NMI}
	\end{equation}
	where, $I(x_{\beta, h}, x_\beta)$ represents the mutual information between $x _{\beta, h}$ and $x_{\beta}$. $H(x_{\beta, h})$ and $H(x_{\beta})$ mean the information entropy of $x_{\beta, h}$ and $x_{\beta}$, respectively. As a result, the distillation loss  can be expressed as
	\begin{equation}
		{{\cal L}_{DL}} = {{\cal L}_{KL}}  + {\cal L}^{T}_{NMI} + {\cal L}^{S}_{NMI},
	\end{equation}
	where ${\cal L}^{T}_{NMI}$ and ${\cal L}^{S}_{NMI}$ represent the NMI loss of the teacher network and the student network, respectively.
	
	\textbf{CE Loss} Cross-entropy (CE) loss, as a widely favored loss function in deep learning, uniquely quantifies the deviation between the predicted probability distribution and the true label distribution, demonstrating remarkable effectiveness in driving model optimization. Therefore, we also introduce it into the loss function to precisely guide the training process. CE loss can be expressed as
	\begin{equation}
		{{\cal L}_{CL}} = {\cal L}_{CE}^T + {\cal L}_{CE}^S,
	\end{equation}
	where ${\cal L}_{CE}^T$ and ${\cal L}_{CE}^S$ represent the cross-entropy loss of the teacher network and the student network, respectively.
	
	As a consequence, the total loss function of our network is expressed as
	\begin{equation}
		{\cal L} = {{\cal L}_{DL}}+{{\cal L}_{CL}},
	\end{equation}
	where ${\cal L}_{DL}$ stands for distillation loss and ${\cal L}_{CL}$ represents cross-entropy loss.
	
	\section{Experiments}
	\label{4}
	This section will delve into the performance of our proposed algorithm on point cloud classification and segmentation tasks, and compare it with several classical rotation-sensitive networks, including PointNet \cite{qi2017pointnet}, PointNet++ \cite{qi2017pointnet++}, PointCNN \cite{li2018pointcnn}, and DGCNN \cite{phan2018dgcnn}, as well as the networks with resistance to rotation perturbation such as SPRIN \cite{you2021prin}, Li $et\ al.$ \cite{li2021closer}, LGRNet \cite{zhao2022rotation}, and Riconv++ \cite{zhang2022riconv++}. Given that the rotation perturbation is an unexpected small change, we apply random angles within [0\degree, 30\degree] to rotate the three-dimensional coordinates of each point cloud within the test set to verify the robustness. It is noteworthy that the networks employing the farthest point sampling (FPS) method \cite{qi2017pointnet++} during inference phase may select different farthest points each time, potentially leading to fluctuations in network performance \cite{lang2020samplenet}. To ensure the fairness of experiments, we set a fixed seed for the networks to guarantee consistency in the selection of farthest points and adopt a three-fold voting strategy to reduce experimental errors. All experiments are conducted on a desktop PC equipped with an NVIDIA GeForce RTX 4090 and 64 GB RAM. During the training process, we employ the Adam optimizer, batch size is 32, and learning rate is 0.001.
	
	\begin{table}
		\renewcommand{\arraystretch}{1.3}
		\setlength{\tabcolsep}{15pt}
		\caption{Classification accuracy (\%) on ModelNet40 and ScanobjectNN after applying the random rotation to the test set. Notably, (aug) refers to networks enhanced by data augmentation technique. The best results are highlighted in boldface and the second best results are underlined.}
		\label{classification}
		\centering
		\begin{tabular}{lll}
			\toprule
			Networks  & ModelNet40 &ScanobjectNN \\ 
			\cmidrule(r){1-3}
			PointNet   & \quad 65.80 & \qquad 68.92 \\
			PointNet++  & \quad 65.45 & \qquad 70.63 \\
			PointCNN  & \quad 60.56& \qquad 67.54 \\ 
			DGCNN   & \quad 80.55 & \qquad 69.29 \\
			PointNet(aug)   & \quad 86.67 & \qquad 75.90 \\
			PointNet++(aug)  & \quad 88.04 & \qquad 76.93 \\
			PointCNN(aug)  & \quad 85.11 &\qquad  75.78 \\ 
			DGCNN(aug)   &  \quad \underline{88.73} & \qquad 78.00 \\
			SPRIN   & \quad 80.03 & \qquad 77.04 \\
			Li $et\ al.$  & \quad 83.74 & \qquad 72.31 \\ 
			LGRNet   & \quad 86.87 & \qquad \underline{80.33} \\ 
			Riconv++  & \quad 88.23 & \qquad 80.01 \\
			Ours  & \quad \textbf{89.63} & \qquad \textbf{81.84} \\
			\bottomrule
		\end{tabular}
	\end{table}

	\subsection{Classification Results}
	\label{4.1}
	\textbf{ModelNet40} ModelNet40 \cite{wu20153d} is a widely utilized dataset for 3D object classification, encompassing 40 distinct categories with 12,311 models, of which 9,843 are designated for training  and 2,468 are reserved for testing.  Notably, considering that data augmentation is a commonly used technique to effectively resist the rotation perturbation, we further present the results obtained by classical point cloud networks when the training sets undergo random rotation within the range [0\degree, 30\degree]. Table \ref{classification} presents the detailed classification accuracy of these networks on ModelNet40 after applying random rotation to the test set, where (aug) represents the classification results of the network after data augmentation.
	
	\begin{table*}
		\renewcommand{\arraystretch}{1.3}
		\setlength{\tabcolsep}{3pt}
		\caption{Average category mIoU(\%) on Shapenet after applying random rotation to the test set. (aug) refers to networks enhanced by data augmentation technique. The best results are highlighted in boldface and the second best results are underlined.}
		\label{partseg}
		\centering
		\begin{tabular}{lccccccccccccccccc}
			\toprule
			Networks  & Avg\_mIoU &  Aero & Bag & Cap & Car & Chair & Earph. & Guitar & Knife & Lamp & Laptop & Motor & Mug & Pistol & Rocket & Skate & Table \\ 
			\cmidrule(r){1-18}
			PointNet & 58.69 & 46.42 & 62.53 & 72.68 & 39.29 & 69.59 & 40.10 & 74.04 & 74.17 & 64.78 & 81.25 & 33.93 & 76.01 & 56.98 & 44.44& 40.79 & 61.93 \\
			PointNet++ & 71.01 & 65.80 & 71.41 & 71.16 & 50.27 & 79.82 & 67.22 & 83.82 & 84.57 & 76.51 & 92.75 & 45.84 & 90.62 & 68.71 & 48.21& 68.46 & 70.87\\
			PointCNN & 64.52 & 60.58 & 78.76 & 61.23 & 42.37 & 70.72 & 71.52 & 76.83 & 77.46 & 65.42 & 76.43 & 38.62 & 77.60 & 55.26 & 46.24 & 61.26 & 72.01\\
			DGCNN & 65.07 & 66.64 & 68.45 & 62.28 & 41.12 & 79.15 & 57.12 & 86.14 & 84.09 & 76.87 & 88.74 & 32.42 & 88.43 & 68.04 & 31.53 & 43.42 & 66.58\\
			PointNet(aug) & 77.35 & 79.72 & 69.46 & \underline{85.00} & 69.44 & 88.70 & 70.96 & 88.14 & 86.52 & 78.85 & 94.90 & 54.65 & 89.72 & 77.34 & 54.23& 68.27 & 81.69 \\
			PointNet++(aug) & 80.82 & 82.57 & 79.00 & 83.52 & 76.62 & 89.27 & 67.76 & 89.81 & \underline{87.06} & \underline{83.44} & 94.52 & 70.13 & 92.02 & \underline{79.00} & \underline{61.59} & 75.11& 82.65\\
			PointCNN(aug) & 76.35 & 81.58 & \textbf{83.76} & 79.23 & 74.37 & 82.72 & 73.52 & 85.83 & 80.46 & 72.42 & 81.43 & \underline{65.62} & 80.02 & 66.26 & 61.24 & 70.25 & \underline{83.01}\\
			DGCNN(aug) & 76.63 & 80.65 & 73.78 & 79.58 & 72.80 & 89.52 & 70.80 & 89.43 & 86.20 & 82.12 & \underline{95.40} & 40.31 & 91.88 & 76.44 & 47.52 & 66.82 & 82.82\\
			SPRIN & 75.30 & 80.31 & 74.12 & 70.57 & 70.02 & 87.78 & 73.00 & 89.93 & 81.75 & 80.89 & 79.98 & 53.41 & 86.60 & 75.95 & 53.87 & 68.61 & 80.00\\
			Li $et\ al.$  & 73.17 & 80.26 & 57.17 & 72.68 & 73.49 & 89.12 & 67.24 & 89.16 & 83.62 & 80.22 & 85.70 & 32.81 & 86.06 & 71.91 & 49.37 & 70.43 & 81.55\\
			LGRNet & 69.23 & 78.99 & 49.55 & 77.74 & 68.44 & 85.42 & 60.84 & 88.28 & 84.10 & 80.24 & 78.52 & 27.60 & 81.39 & 55.58 & 46.74 & 64.72 & 79.42\\
			Riconv++ & \underline{81.28} & \textbf{84.74} & \underline{81.65} & \textbf{87.36} & \underline{78.13} & \textbf{90.55} & \underline{75.79} & \textbf{91.36} & 81.60 & 81.99 & 94.50 & \textbf{65.99} & \underline{94.32} & 74.47 & 59.63 & \underline{76.07} & 82.32\\
			Ours & \textbf{82.10} & \underline{83.08} & 80.60 & 83.37 & \textbf{78.81} & \underline{90.34} & \textbf{77.65} & \underline{90.96} & \textbf{87.60} & \textbf{84.09} & \textbf{96.40} & 60.66 & \textbf{95.31} & \textbf{80.73} & \textbf{63.91} & \textbf{76.58} & \textbf{83.46}\\
			\bottomrule
		\end{tabular}
	\end{table*}
	
	From Table \ref{classification}, we can observe that the classification accuracy of classical rotation-sensitive networks is unsatisfactory when unforeseen rotations occur in the test set. Considering that during the process of point cloud acquisition, equipment offset and system instability may lead to biases in the data, it is evident that classical networks easily struggle in such scenarios, while data augmentation and rotation-robust networks are better choices. Among rotation-robust networks, SPRIN and Li $et\ al.$ achieve the rotation robustness through spherical voxel convolution and PCA method, respectively. Although these networks nearly realize consistency in different angles, the classification accuracy is far below the ideal level. LGRNet and Riconv++ convert three-dimensional coordinate information into multi-dimensional invariant features such as distance and angle, improving accuracy but also introducing redundant calculations. In contrast, the data augmentation technique can improve the rotation robustness, but it is unsatisfactory in classification accuracy due to the inherent structure of the network and unbalanced training distribution. Different from them, our proposed distillation method not only directly utilizes the original three-dimensional coordinates during the inference phase, but also maintains optimal performance under rotation perturbations, as evidenced by an average increase in accuracy of 17.28\% compared to classic rotation-sensitive networks, and an improvement of 4.92\% and 2.50\% respectively compared to rotation-robust networks and data augmentation-based networks.
	
	\textbf{ScanobjectNN} ScanObjectNN\cite{uy2019revisiting} is an indoor scene scanning dataset which contains 15,000 objects, categorized into 15 classes with 2902 unique object instances. Compared to the regular classification dataset ModelNet40, the ScanobjectNN dataset is more challenging because it contains a large amount of missing and distorted data. Table \ref{classification} also lists the classification accuracy of these networks on ScanobjectNN. The following observations can be obtained from Table \ref{classification}. Firstly, due to the numerous distorted and missing data in ScanobjectNN, the overall accuracy is significantly lower than that on ModelNet40, indicating that current networks still have certain limitations when recognizing complex tasks. Secondly, the classification accuracy of classical rotation-sensitive classification networks is notably lower than that of rotation-robust networks and data augmentation-based networks, which also verifies the previously mentioned conclusion that classical rotation-sensitive point cloud networks struggle in the case of data deviations. Thirdly, although rotation-robust networks and data augmentation-based networks have a certain degree of resistance to the rotation perturbation, there is still room for improvements in terms of classification accuracy. In contrast, our network not only achieves the perturbation robustness without coordinate transformation but also achieves better results than other networks. Specifically, our accuracy improves by an average of 12.74\% compared to classic rotation-sensitive networks, and by an average of 4.41\% and 5.19\% respectively compared to rotation-robust networks and data augmentation-based networks.

	\subsection{Segmentation Results}
	
	\textbf{ShapeNet} Compared with the classification task, the segmentation task is more challenging because it involves the recognition of fine-grained structures and intricate spatial relationships between different object parts. We here conduct experiments on ShapeNet \cite{chang2015shapenet}, one widely used part segmentation dataset for 3D shape understanding and analysis, encompassing 16,881 shapes from 16 object categories with 50 part labels in total. The average category mIoU \cite{zhao2022rotation} is utilized to compare the segmentation performance, which is presented in Table \ref{partseg}. As evident from Table \ref{partseg}, when experiencing unpredictable rotations in the test set, the segmentation performance of classical rotation-sensitive point cloud networks is generally poor. However, upon application of the data augmentation method, marked improvements in the average category mIoU across various networks are observed, conclusively demonstrating that data augmentation effectively enhances the networks' resilience against the rotation perturbation. In addition, although some networks inherently possess the rotation robustness, exhibiting good tolerance to angular variations, the majority still fails in terms of satisfactory performance. Notably, Riconv++ demonstrates formidable capabilities in handling complex 3D shapes, yet its computational speed and noise tolerance necessitate further optimization to achieve an optimal state, which can be proved by the following experiments. In contrast, our proposed method achieves an average improvement of 17.28\% in mIoU compared to classic rotation-sensitive networks, while demonstrating an average increase of 7.36\% and 4.32\% in mIoU respectively over rotation-robust networks and data augmentation-based networks. 
	
	\textbf{S3DIS} S3DIS \cite{armeni20163d}, serving as a pivotal dataset for 3D semantic segmentation in the field of computer vision, encompasses 6 indoor scenes, totaling 271 detailed room data. Due to the intricacy of indoor environments, such as high occlusion between objects, varying lighting conditions as well as intricate and variable layout structures, the semantic segmentation task posed by S3DIS is particularly challenging. We present the experiment results in Table \ref{semseg}. 
	Given the high complexity of scenarios, classical rotation-sensitive point cloud networks exhibit significant limitations when rotation perturbations are introduced into the test set, with segmentation accuracy far below that of other networks, making it nearly impossible to accurately distinguish between different scenes. After applying the data augmentation, the performance of these networks has been significantly improved, indicating that data augmentation is beneficial in reducing the network's sensitivity to the rotation perturbations. At the same time, rotation-robust networks also show superior resistance to the rotational interference. However, the accuracy of classical rotation-sensitive networks based on data augmentation, as well as that of the majority of rotation-robust networks, still fall short of the ideal level, suggesting that these networks struggle to recognize point cloud in complex scenarios. It is noteworthy that our method achieves an average improvement of 40.16\% and 12.41\% in mIoU compared to rotation-sensitive networks and data augmentation-based networks, respectively. Additionally, our method also demonstrates an average increase of 4.82\% compared to rotation-robust networks, except for Riconv++. Even so, compared to Riconv++, our method can significantly improve noise and outlier tolerance (which will be verified by \ref{Robustness_noise} and \ref{speed}), indicating that our approach still holds an immense potential for optimization when dealing with complex scenes.
	
	\begin{table}
		\renewcommand{\arraystretch}{1.3}
		\setlength{\tabcolsep}{15pt}
		\caption{Average mIoU (\%) on S3DIS after applying the random rotation to the test set. (aug) refers to networks enhanced by data augmentation technique. The best results are highlighted in boldface and the second best results are underlined.}
		\label{semseg}
		\centering
		\begin{tabular}{lc}
			\toprule
			Networks  & Avg\_mIoU \\ 
			\cmidrule(r){1-2}
			PointNet & 11.76\\
			PointNet++ & 14.75\\
			PointCNN & 13.52\\
			DGCNN & 14.97\\
			PointNet(aug) & 37.59\\
			PointNet++(aug) & 46.81\\
			PointCNN(aug) & 40.23 \\
			DGCNN(aug) & 41.36\\
			SPRIN & 48.32\\
			Li $et\ al.$  & 45.04\\
			LGRNet & 46.17\\
			Riconv++ & \textbf{56.83}\\
			Ours & \underline{53.91}\\
			\bottomrule
		\end{tabular}
	\end{table}

	\subsection{Visualization Analysis}
	
	\begin{figure*}
		\centering
		\includegraphics[width=7in]{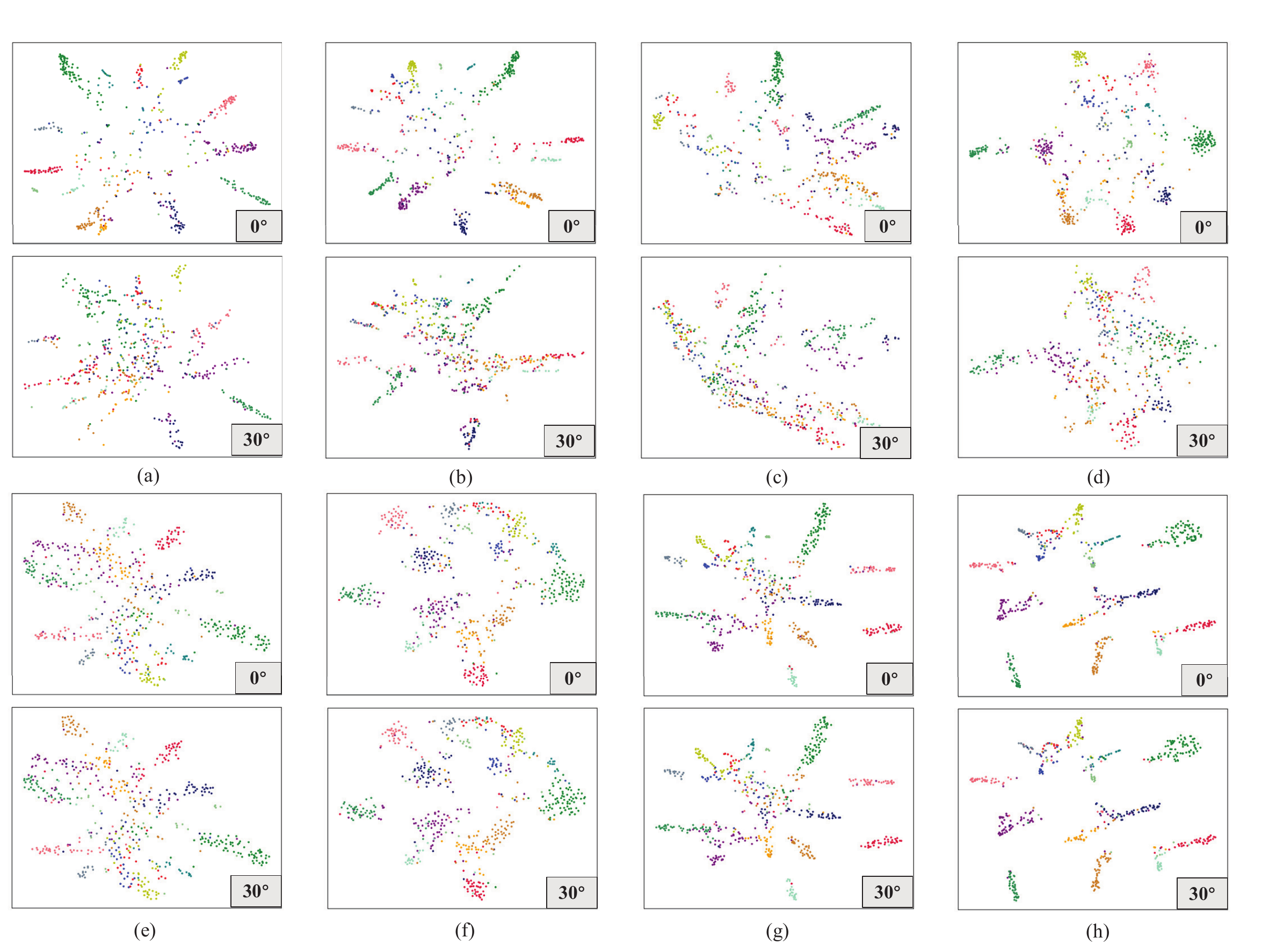}
		\caption{The t-SNE results of different networks at 0\degree and 30\degree, where 0\degree/30\degree refer to the corresponding degrees of rotation on coordinate axis. Different colors represent different categories. (a) PointNet; (b) PointNet++; (c) PointCNN; (d) DGCNN; (e) SPRIN; (f) Li $et\ al$; (g) Riconv++; (h) Ours.}
		\label{visual}
	\end{figure*}
	\begin{figure*}
		\centering
		\includegraphics[width=7in]{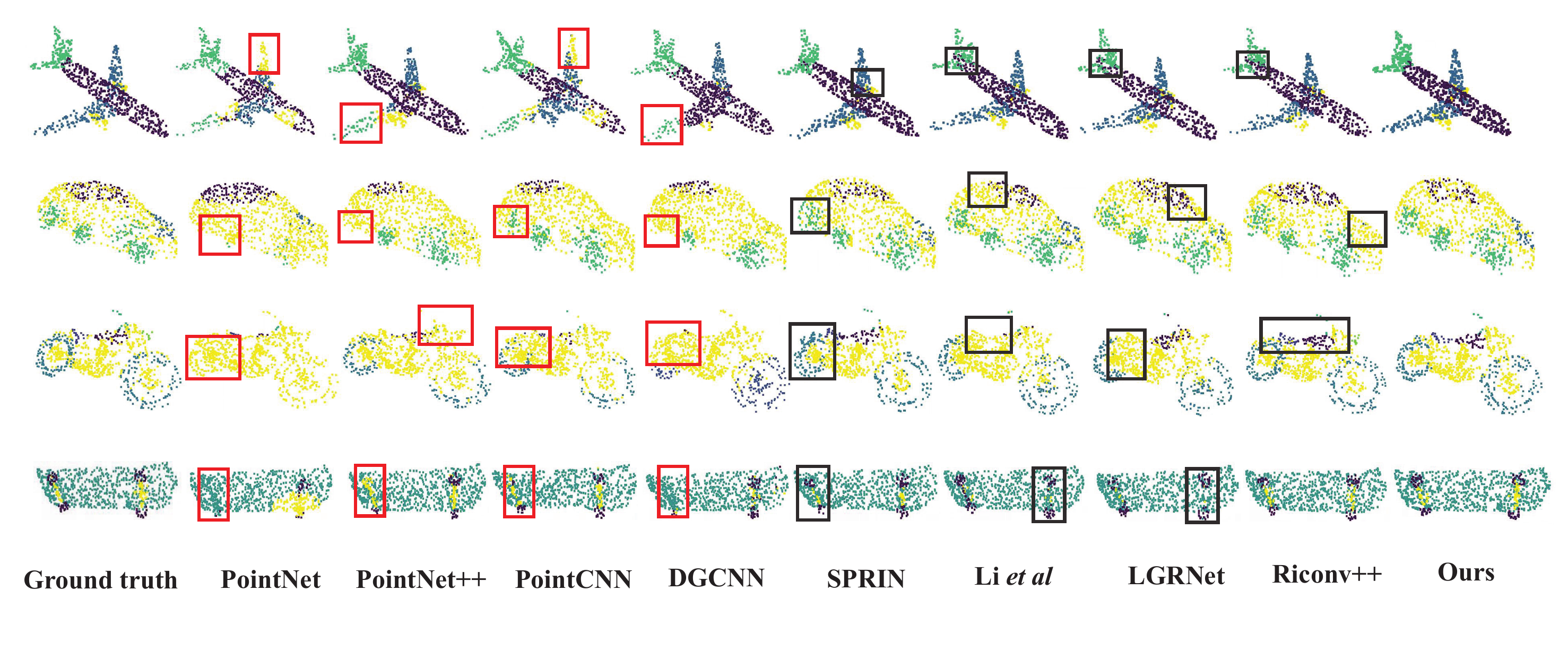}
		\caption{The visualization results for part segmentation. The networks are trained and tested on ShapeNet, and different parts are represented by different colors. Notably, the test data are transformed by the specific 3D rotation (20\degree, 20\degree, 20\degree).}
		\label{visual_part}
	\end{figure*}
	
	\begin{figure*}
		\centering
		\includegraphics[width=6in]{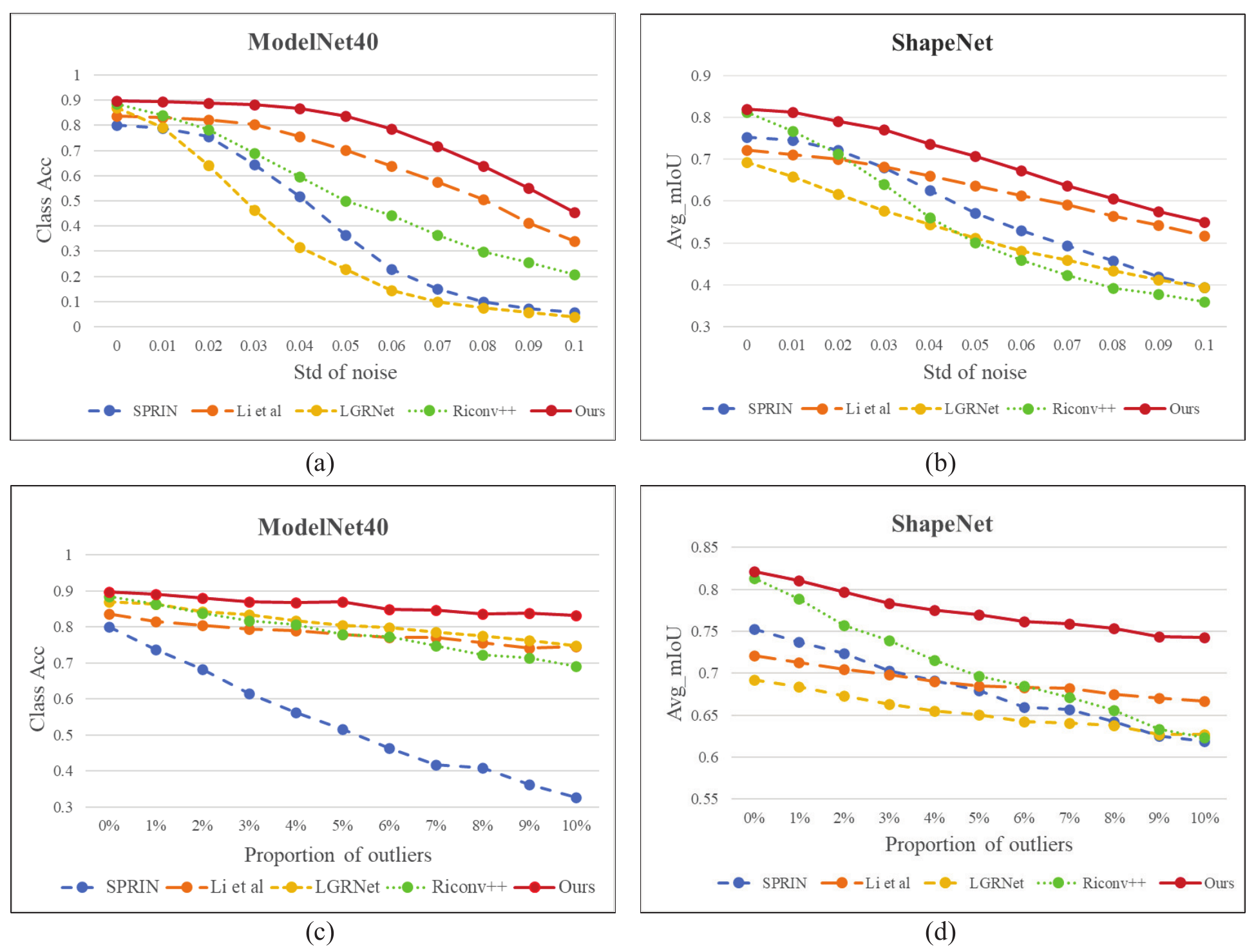}
		\caption{The robustness of different networks to noise and outliers on ModelNet40 and ShapeNet. (a) Robustness to noise on ModelNet40; (b) Robustness to noise on ShapeNet; (c) Robustness to outliers on ModelNet40; (d) Robustness to outliers on ShapeNet. In (a) and (b), the x-axis represents the standard deviation of the noise, the y-axis stands for classification accuracy, while in (c) and (d), the x-axis respectively represents the proportion of outlier points and the average category mIoU.}
		\label{noise}
	\end{figure*}
	
	To intuitively demonstrate the effectiveness of our method, we employ the t-SNE \cite{van2008visualizing} technique for the visualization analysis of ScanobjectNN, concurrently showcasing the visualization results of part segmentation dataset ShapeNet. 
	
	Specifically, when applying the t-SNE algorithm, we reduce the logits of the penultimate layer of networks from 512 dimensions to 2 dimensions, and present visualization results of 0\degree and 30\degree in Fig. \ref{visual}, where different colors represent different categories. We emphasize if the data points on t-SNE form distinct and compact clusters, with clear boundaries between different clusters, it will serve as a strong indicator of good classification performance. As can be clearly observed from Fig. \ref{visual}, when point cloud data are in its initial 0\degree orientation, classical rotation-sensitive networks exhibit distinct inter-class distances, indicating excellent classification performance. Notably, the intra-class variances of PointNet and PointNet++ are smaller than PointCNN and DGCNN, demonstrating superior performance. However, when the data points are rotated by 30\degree, the distribution of data points in the classical rotation-sensitive networks becomes relatively disorganized, which reveals a significant drop in classification performance. In contrast, as seen in Fig. \ref{visual}, the networks with rotation robustness maintain good consistency at 0\degree and 30\degree. Among them, SPRIN and Riconv++ exhibit compact intra-class data points but with less distinct inter-class differences, while Li $et\ al.$ displays clear inter-class gaps yet larger intra-class variations.
	In comparison, our network not only exhibits excellent consistency in the distribution of data points but also achieves sufficient compactness within classes and dispersion between classes, demonstrating excellent classification performance.
	
	Fig. \ref{visual_part} displays some typical visualization results of part segmentation on ShapeNet, comparing the segmentation results of different networks when the test data are transformed by the specific 3D rotation (20\degree, 20\degree, 20\degree). It is still evident that classical rotation-sensitive point cloud networks struggle to maintain stable segmentation performance after data rotation, as evidenced by the numerous misclassified points in the segmentation results highlighted by the red boxes in Fig. \ref{visual_part}. This directly reflects their inability to cope with the challenges posed by rotation perturbation. While the networks with rotation robustness can generally restore the shape of objects and effectively resist rotational interference, their segmentation performance remains inadequate when dealing with finer details, such as the tail details of an airplane and the wheels of a motorcycle, as indicated by the black boxes in Fig. \ref{visual_part}. Contrastively, our method not only excels in rotation robustness but also demonstrates outstanding performance in segmenting intricate details.
	
	\subsection{Robustness to Noise and Outliers}
	\label{Robustness_noise}
	In the process of point cloud data acquisition, in addition to the rotation perturbation, random errors such as noise and outliers may also cause interference. However, existing research indicates that most rotation-robust networks tend to perform poorly when confronted with noise and outlier interference \cite{kim2020rotation}. Therefore, we conduct extensive experiments on ModelNet40 and ShapeNet aimed at comparing the effectiveness of our proposed method with current rotation-robust networks in terms of their resistance to noise and outliers.
	
	To validate the robustness of networks to noise, we introduce Gaussian noise with a standard deviation ranging from 0 to 0.1 on each input point. Fig. \ref{noise}(a) presents the experimental results on ModelNet40. It can be found that existing rotation-robust networks tend to experience a faster performance decline as noise levels rise, while our network exhibits a slower performance degradation trend, highlighting superior resistance to noise. Fig. \ref{noise}(b) presents the experimental results on ShapeNet. It can be observed that with the gradual increase of noise level, the performances of SPRIN, LRGNet, and Riconv++ rapidly decline. Although the network proposed by Li $et\ al.$ decreases gently, its overall accuracy does not reach the desired level. Different from them, our network holds superior segmentation performance under various noise conditions as well as possesses the strong noise resistance capabilities.
	
	To illustrate the impact of outliers on network performance, we randomly select 1\% to 10\% of the sampling points for each object and add noise with the standard deviation of 0.1 to simulate data pollution. Fig. \ref{noise}(c) presents the experimental comparison on ModelNet40. As the proportion of outliers increases, the performance of SPRIN significantly declines, while the remaining four networks demonstrate superior resistance to outlier. Notably, our network stands out in these networks, exhibiting the best performance. Fig. \ref{noise}(d) further reveals the experimental results on ShapeNet. It can be seen that the performance of Riconv++ noticeably degrades as the proportion of outliers increases. Although networks such as SPRIN, Li $et\ al.$, and LGRNet show commendable resistance to outliers, they fall short in terms of segmentation accuracy. In contrast, our network maintains a strong advantage in outlier resistance as well as surpasses all other networks by a wide margin in accuracy, demonstrating its comprehensive superiority.
	
	\subsection{Analysis of Inference Speed}
	\label{speed}
	\begin{figure}
		\centering
		\subfloat[]{
			\includegraphics[scale=0.5]{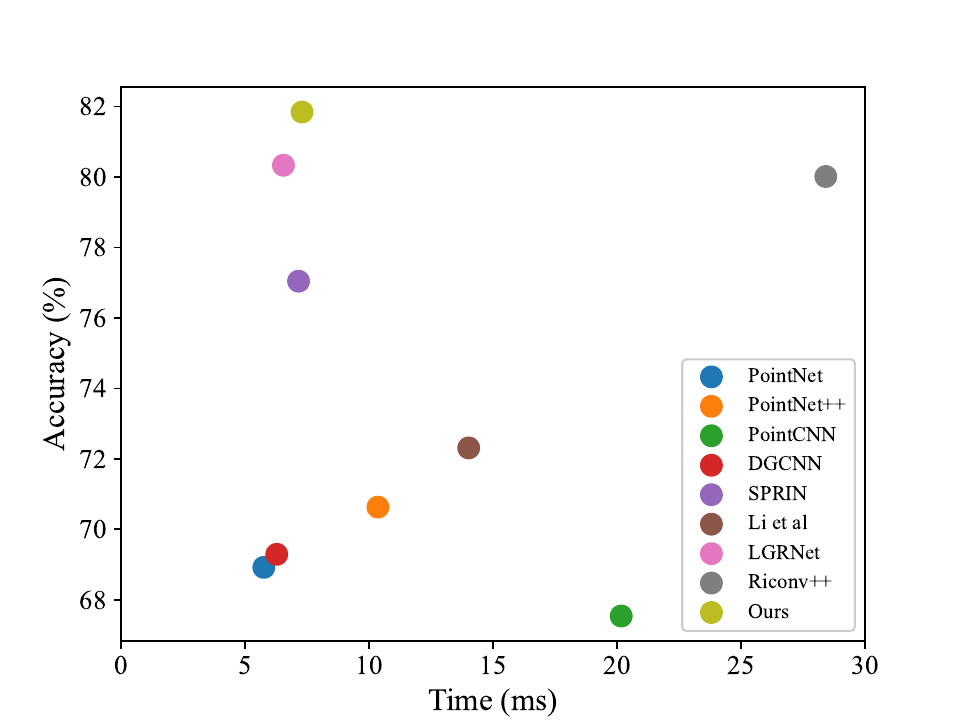}}\\[-0.5mm]
		\subfloat[]{
			\includegraphics[scale=0.5]{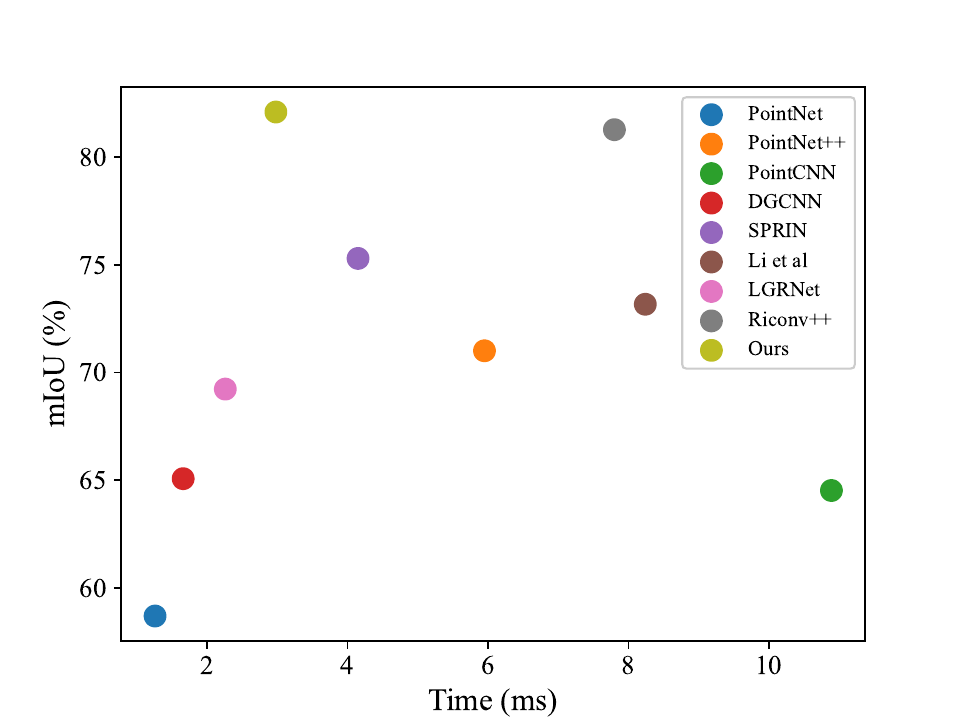}}
		\caption{The speed and performance of different networks on ScanobjectNN and ShapeNet. (a) ScanobjectNN; (b) ShapeNet.}
		\label{Speed}
	\end{figure}
	
	To evaluate the networks' performance in terms of inference time, taking the datasets ScanobjectNN and ShapeNet for example, we further test their operation speeds. Specifically, the speed is tested on 4090 with a batch size of 32, full precision (fp32), and measured in seconds/examples. The experimental results are shown in Fig. \ref{Speed}. As is evident from Fig. \ref{Speed}(a), the inference speeds of Li et al., PointCNN, and Riconv++ are significantly slower than ideal expectations, while PointNet, DGCNN, and PointNet++ exhibit considerable advantages in terms of inference speed, but their classification accuracies are markedly lower than those of rotation-robust networks. In comparison, while LGRNet and SPRIN exhibit similar speeds to our network, our network achieves a superior balance between accuracy and speed. Similarly, as can be seen from Fig. \ref{Speed}(b), most networks are inferior to ours in segmentation performance. Although Riconv++ can achieve comparable segmentation performance with our network, the speed of it is much slower than ours. As a result, our method can obtain a more satisfactory trade-off between performance and time consumption than the other SOTA methods.

	\subsection{Ablation Experiments}
	To comprehensively validate the effectiveness of our proposed MM model and alignment model, we have further conducted ablation experiments on the highly challenging classification dataset ScanobjectNN to investigate the classification accuracy of using solely the student network (S), the teacher-student architecture (T-S), the teacher-student architecture fused with attention maps (T-S+Attn), and the final alignment model integrated with normalized mutual information (T-S+Attn+NMI). The experimental results have been listed in Table \ref{Ablation}.
	
	\begin{table}
		\renewcommand{\arraystretch}{1.3}
		\setlength{\tabcolsep}{15pt}
		\caption{Classification accuracy (\%) on ScanobjectNN under only student network (S), the teacher-student architecture (T-S), the teacher-student architecture fused with attention maps (T-S+Attn), and the teacher-student architecture with the final alignment model (T-S+Attn+NMI).}
		\label{Ablation}
		\centering
		\begin{tabular}{lc}
			\toprule
			Networks  & Accuracy \\ 
			\cmidrule(r){1-2}
			S & 71.83 \\
			T-S & 69.24 \\
			T-S+Attn & 80.95 \\
			T-S+Attn+NMI & 81.84 \\
			\bottomrule
		\end{tabular}
	\end{table}
	
	It is apparent from Table \ref{Ablation} that when relying solely on the student network for classification prediction, due to its lack of design for rotation robustness, the classification accuracy of network drops significantly once the test set data undergo random rotation, far below the ideal level. More notably, if the features of the teacher network and the student network are directly aligned without any special design, the loss of classification accuracy will be even more severe. This is primarily attributed to the significant differences in the feature spaces of the two networks, where direct information transfer can lead to the distortion and loss of critical information, thereby severely impacting the accuracy of object recognition. However, by introducing the attention map from the designed alignment model (i.e., Eq. (\ref{loss})), we successfully overcome this challenge and achieve a substantial improvement in performance. Furthermore, through incorporating normalization information to constrain the network (i.e., Eq. (\ref{NMI})), we have ultimately obtained the most outstanding classification results. These experiments fully demonstrate the effectiveness of our proposed algorithm in enhancing model performance.
	
	\section{Limitations}
	\label{5}
	In essence, our distillation method transfers the information from the teacher network to the student network for altering the search path in the solution space of network. However, this method can only induce the output results of network to approach the fitting limit, unable to break through the network's design architecture to achieve full-pose rotation invariance. Therefore, the current augmentation method based on manifold distillation can only be applied to a certain range of angles to resist equipment deviation, system instability or calculation errors and so on. Our future research will attempt to change the embedding pose of the manifold to achieve perturbation robustness over a wider range of angles.
	
	\section{Conclusion}
	\label{6}
	This paper introduced an augmentation strategy based on manifold distillation technology, aiming to enhance the robustness against the rotation perturbation. This strategy adopted a teacher-student network architecture, where the teacher network focused on learning the rotation-robust information during the training phase and transferred this information to the student network through online distillation. Furthermore, we had designed the alignment model to ensure that the manifold structure of student network can be precisely aligned with that of the teacher network, thereby guaranteeing the effectiveness of information transfer during the distillation process. As a result, during the inference phase, the student network can directly utilize the original 3D coordinate data to achieve rotation perturbation robustness without any coordinate transformation. In addition, experiments on existing point cloud classification and segmentation datasets showed that our method not only demonstrates the robust performance in resisting rotation perturbation but also exhibits the exceptional resilience against the noise and outliers.

	
	%



	\ifCLASSOPTIONcaptionsoff
	\newpage
	\fi

	
	
	%
	
	{
		\small
		\bibliographystyle{unsrt}
		\bibliography{ref}
	}
	%
	\begin{IEEEbiography}[{\includegraphics[width=1in,height=1.25in,clip,keepaspectratio]{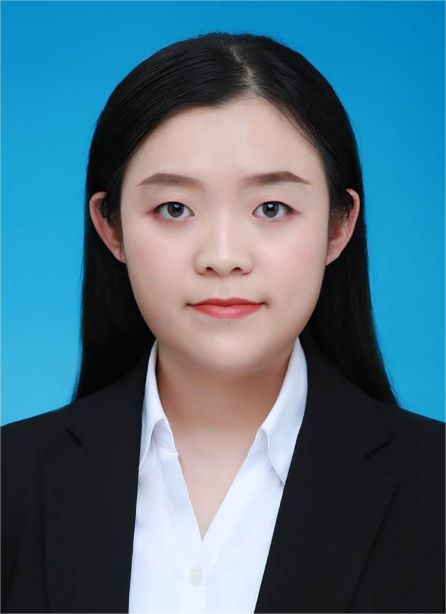}}]{{Xinyu Xu}}
		received the B.E. degree. in measurement and control technology and instrument from Tianjin University, Tianjin, China in 2022, and she is currently pursuing the M.Sc. degree in electronic information from Shanghai Jiao Tong University, Shanghai, China. 
		
		Her research interests include computer vision, point cloud recognition, and deep learning in artificial intelligence.
	\end{IEEEbiography}
	
		\begin{IEEEbiography}[{\includegraphics[width=1in,height=1.25in,clip,keepaspectratio]{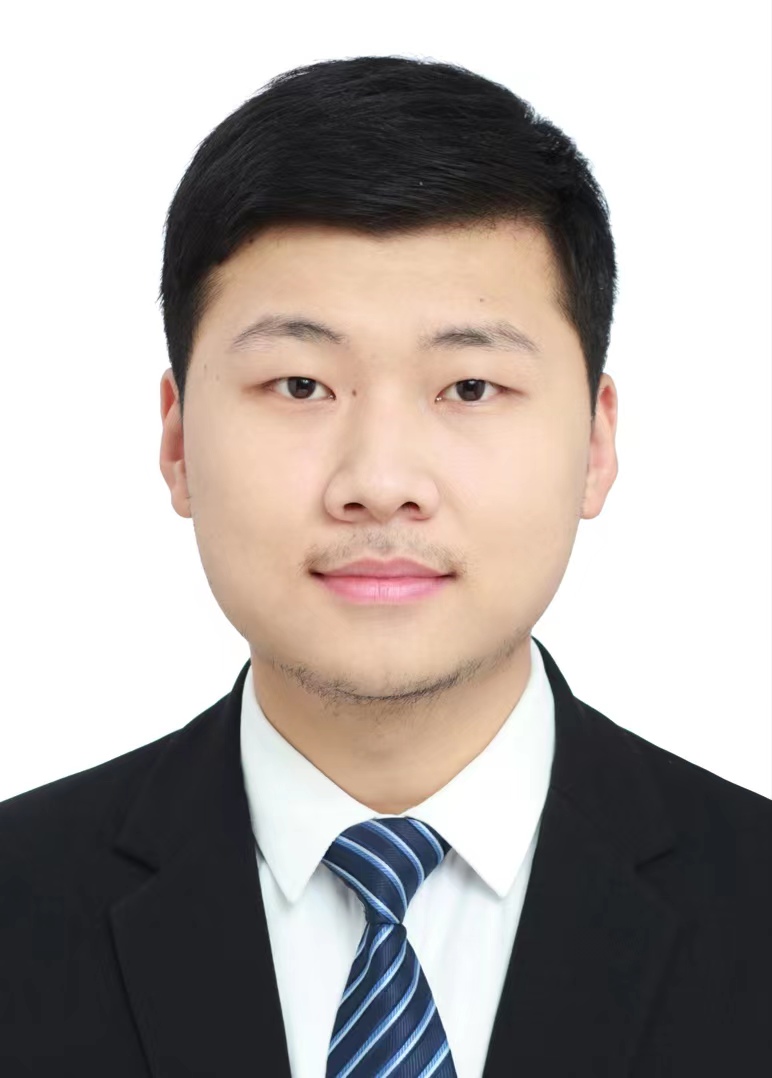}}]{{Huazhen Liu}}
		received the B.E. degree in measurement and control technology and instrument from Tian Jin University Tianjin,China, in 2022, and he is currently pursuing the M.Sc. degree in Shanghai Jiao Tong University, Shanghai, China. 
		
		His primary research interests include computer vision and optical engineering in artificial intelligence.
	\end{IEEEbiography}

	\begin{IEEEbiography}[{\includegraphics[width=1in,height=1.25in,clip]{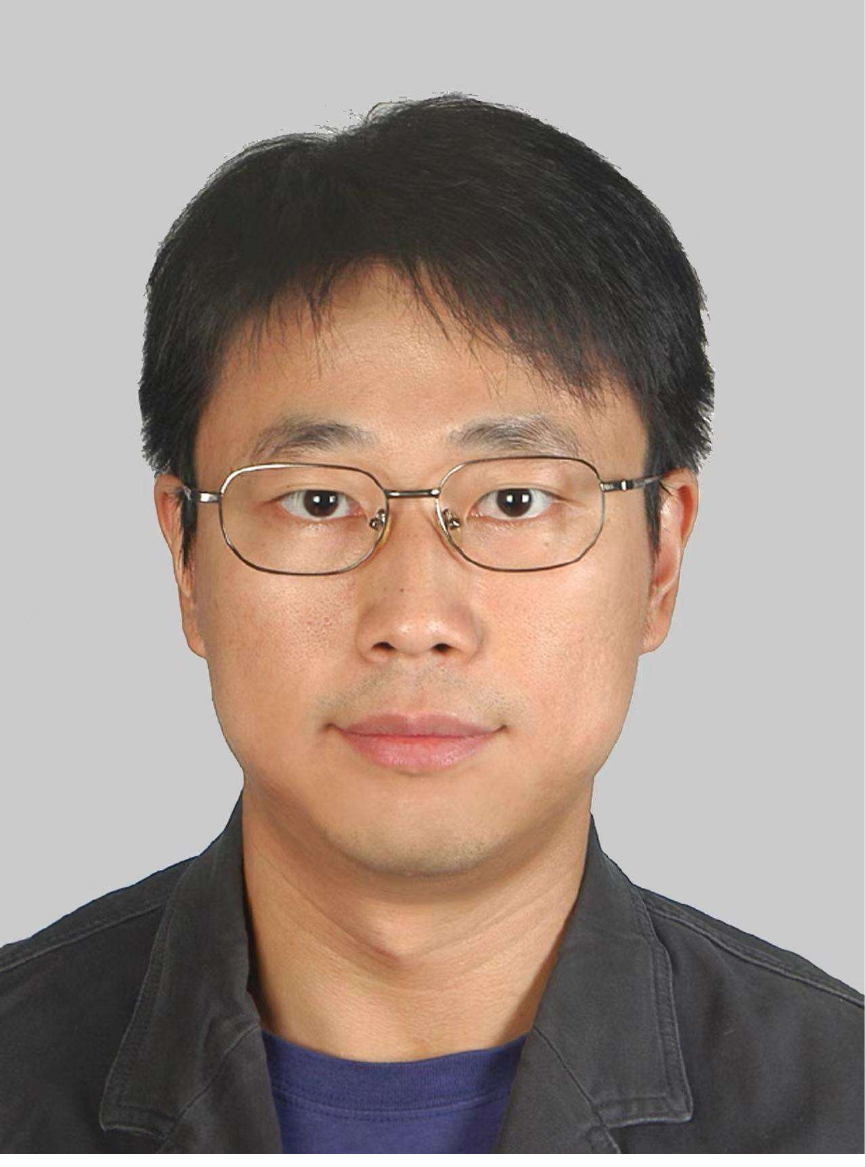}}]{Feiming Wei} received the Ph.D degree in electronic information from Shanghai Jiao Tong University, China, 2021.
		
		He is currently working as a Professor in the Shanghai Key Laboratory of Intelligent Sensing and Recognition, Shanghai Jiao Tong University. His research work focuses on the characteristics of radar target and intelligent object recognition.
	\end{IEEEbiography}
	
	\begin{IEEEbiography}[{\includegraphics[width=1in,height=1.25in,clip]{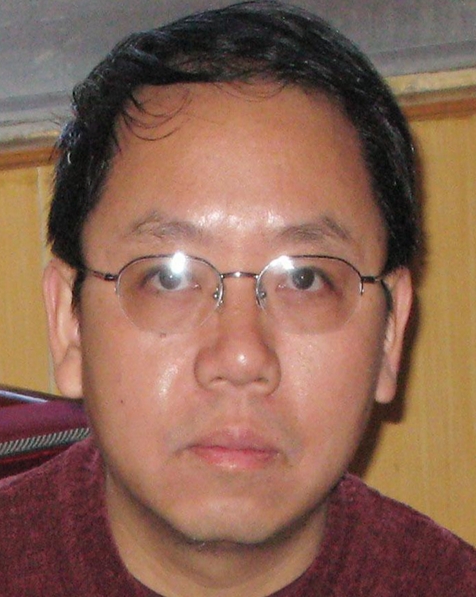}}]{Huilin Xiong} received the B.Sc. and M.Sc. degrees in mathematics from Wuhan University, Wuhan, China, in 1986 and 1989, respectively, and the Ph.D. degree in pattern recognition and intelligent control from the Institute of Pattern Recognition and Artificial Intelligence, Huazhong University of Science and Technology, Wuhan, in 1999. He joined Shanghai Jiao Tong University, Shanghai, China, in 2007, where he is a Professor with the Department of Automation. 
		
		His research interests include pattern recognition, machine learning, synthetic aperture radar data processing, and bioinfomatics.
	\end{IEEEbiography}
	
	\begin{IEEEbiography}[{\includegraphics[width=1in,height=1.25in,clip,keepaspectratio]{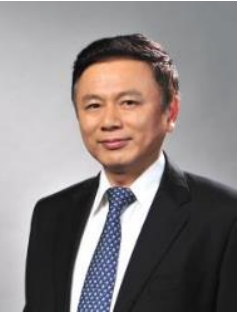}}]{Wenxian Yu} (Senior Member IEEE) was born in Shanghai, China, in 1964. He received the B.Sc., M.Sc., and Ph.D. degrees from the National University of Defense Technology, Changsha, China, in 1985, 1988, and 1993, respectively.
		
		From 1996 to 2008, he was a Professor with the College of Electronic Science and Engineering, National University of Defense Technology, where he served as the deputy dean of the school and assistant director of the National Key Laboratory of Automatic Target Recognition. In 2008, he joined the School of Electronics, Information, and Electrical Engineering, Shanghai Jiao Tong University, Shanghai, in which he served as the executive dean from 2009 to 2011. Currently, he is a Distinguished Professor under the Yangtze River Scholar Scheme, vice dean of the Advanced Industrial Technology Research Institute, and dean of the Academy of Information Technology and Electrical Engineering. His current research interests include radar target recognition, remote sensing information processing, multisensor data fusion, integrated navigation system, etc. In these areas, he has published more than 260 research papers.
	\end{IEEEbiography}
	
	\begin{IEEEbiography}[{\includegraphics[width=1in,height=1.25in,clip,keepaspectratio]{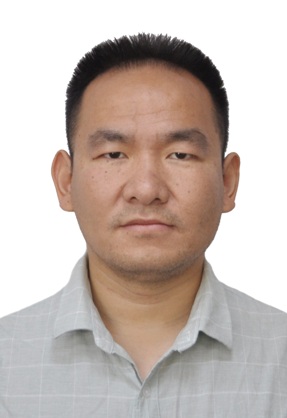}}]{Tao Zhang} (Member IEEE) received the B.S. degree in electronic information engineering from Huainan Normal University in 2011, and the M.Sc. degree in communication and information system from Sichuan University in 2014, and the Ph.D degree in control science and engineering from Shanghai Jiao Tong University in 2019. From 2017 to 2018, he was with the Department of Geoinformatics, the KTH Royal Institute of Technology, as a joint PhD student. From 2019 to 2021, he worked as a Post-Doctoral in the Department of Electronics, Tsinghua University.
		
		He is currently working as an Assistant Professor in the Shanghai Key Laboratory of Intelligent Sensing and Recognition, Shanghai Jiao Tong University. His research work focuses on PolSAR/SAR image processing, object recognition, and machine learning.
	\end{IEEEbiography}
	
\end{document}